\documentclass[times,twocolumn,final,authoryear]{elsarticle}

\usepackage{prletters}
\usepackage{framed,multirow}

\usepackage{amssymb}
\usepackage{latexsym}

\usepackage{url}
\usepackage{xcolor}
\definecolor{newcolor}{rgb}{.8,.349,.1}

\usepackage{adjustbox}
\usepackage{listings}
\usepackage{caption,setspace}
\usepackage{diagbox} 
\usepackage{booktabs} 
\usepackage{todonotes}
\usepackage{svg}
\usepackage[inline,shortlabels]{enumitem}
\usepackage{threeparttable}
\usepackage[perpage]{footmisc}
\newcommand{\eg}{e.g., }

\newcommand{\ie}{i.e., }

\newcommand{\R}{\mathbb{R}}

\usepackage{todonotes}

\usepackage{color}

\newcommand{\figref}[1]{Fig.~\ref{#1}}    
\newcommand{\Figref}[1]{Figure~\ref{#1}}  
\newcommand{\tabref}[1]{Table~\ref{#1}}
\newcommand{\Tabref}[1]{Table~\ref{#1}}
\newcommand{\secref}[1]{Section~\ref{#1}}
\newcommand{\equref}[1]{Eq.~(\ref{#1})}

\usepackage{amssymb}
\usepackage{amsmath}
\usepackage{siunitx}
\newcommand{\h}{\boldsymbol{h}}

\newcommand{\topk}{S}
\newcommand{\FF}{\text{FF}}
\DeclareMathOperator{\softmax}{softmax}

\journal{Pattern Recognition Letters}

\begin{document}

\begin{frontmatter}

\title{Block-wise Dynamic Sparseness}

\author[1]{Amir \snm{Hadifar}\corref{cor1}} 
\ead{amir.hadifar@ugent.be}
\author[1]{Johannes \snm{Deleu}}
\author[1]{Chris \snm{Develder}}
\author[1]{Thomas \snm{Demeester}}

\address[1]{Ghent University -- imec, Department of Information Technology, Technologiepark 126, 9052 Zwijnaarde, Belgium}

\received{1 May 2013}
\finalform{10 May 2013}
\accepted{13 May 2013}
\availableonline{15 May 2013}
\communicated{S. Sarkar}

\begin{abstract}
Neural networks have achieved state of the art performance across a wide variety of machine learning tasks, often with large and computation-heavy models.
Inducing sparseness as a way to reduce the memory and computation footprint of these models has seen significant research attention 
in recent years.
In this paper, we present a new method for \emph{dynamic sparseness}, whereby part of the computations are omitted dynamically, based on the input. 
For efficiency, we combined the idea of dynamic sparseness with block-wise matrix-vector multiplications.
In contrast to static sparseness, which permanently zeroes out selected positions in weight matrices, our method preserves the full network capabilities by potentially accessing any trained weights. Yet, matrix vector multiplications are accelerated by omitting a pre-defined fraction of weight blocks from the matrix, based on the input.   
Experimental results on the task of language modeling, using recurrent and quasi-recurrent models, show that the proposed method can outperform a magnitude-based static sparseness baseline.
In addition, our method achieves similar language modeling perplexities as the dense baseline, at half the computational cost at inference time. 
  
\end{abstract}


\end{frontmatter}

\section{Introduction} \label{sec:introduction}
Deep Neural Networks (DNNs) have been a success story in recent years, due to their impressive performance on various domains. 
Theoretically and empirically, it has been shown that DNNs, trained by first-order-methods such as stochastic gradient decent (SGD), are able to represent a wide variety of complex functions 
\citep{hornik1989multilayer}.
Although in general it seems that
utilizing large and fully parameterized networks is a
reasonable
way to increase 
prediction effectiveness, the computational complexity and memory demand of these models may become a bottleneck.

Many methods have been proposed to address that general issue of computational and memory complexity associated with large models. 
Researchers have focused on carefully redesigning \emph{neural network architectures} to reduce the computational cost while maintaining effectiveness.
Such approaches include:
\begin{enumerate*}[(i)]
\item \emph{tensor decomposition} to express tensors in terms of a sequence of operations on simpler (e.g., smaller) tensors  
\citep{denton2014exploiting},
\item \emph{quantization} to reduce the precision of weights and activation functions with minimal impact on performance \citep{courbariaux2016binarized},
\item \emph{knowledge distillation} to transfer knowledge of a larger model into another lightweight model 
\citep{hinton2015distilling}, 
and 
\item \emph{network pruning} to remove redundant and uncritical connections in order to arrive at sparse models \citep{han2015deep}.  
\end{enumerate*}

This paper can be situated in the area of network pruning. However, whereas most previous work focuses on what we call \emph{static} pruning, \ie permanently disabling specific network connections, we focus on \emph{dynamic} pruning, whereby a well-chosen pruned version of the network is used, depending on the input.

Yet, to avoid inefficiency which would result from a fully flexible sparsity pattern for each input instance, we propose block-wise pruning: entire blocks of model parameters are jointly pruned, thus facilitating an efficient implementation with a limited number of additional parameters to compute the dynamic pruning mask.

We argue that our proposal of dynamic block-wise pruning can achieve a (reduced) computational cost similar to static pruning, while largely keeping the expressiveness of the non-sparse case, in the sense that overall (\ie across all instances) the same number of network parameters can be tuned.
The next section provides an extensive overview of related work in the area of pruning techniques for neural networks.
The subsequent sections further detail our contributions, which can be summarized as follows:
\begin{enumerate*}[(i)]
\item We propose a dynamic block-wise pruning approach (\secref{sec:background}), aimed at achieving low computational cost without affecting expressivity much, 
\item We experimentally validate our approach, illustrating its dynamic gating mechanism on MNIST (\secref{sec:experiment:MNIST}), as well as 
\item the effectiveness of dynamic sparseness in limiting computational complexity while maintaining performance for language modeling (\secref{sec:experiments:LM}).
\end{enumerate*}
The final \secref{sec:conclusion} summarizes our conclusions and indicates potential areas of follow-up research.
  
\section{Related Work}
\label{sec:background}
This section provides an overview of related literature.
We start by providing a high-level overview on various strategies for pruning, 
thereby attempting to bring clarity in the diverse terminology used in literature 
(\secref{subsec:background-pruning-strategy}). We then explore the following sub-domains in more detail:
unstructured pruning (\secref{subsec:unstructured}), structured pruning (\secref{subsec:structured}), and conditional computation (\secref{subsec:conditional}).

\subsection{Overview of Pruning Strategies}
\label{subsec:background-pruning-strategy}

The first notion of sparsity 
in the area of artificial neural networks appeared as pruning techniques \citep{23864} designed to
gradually switch off increasing numbers of 
 network parameters during training, until either performance starts to drop, or a certain level of sparseness is obtained.
It has been found to be a practical solution to reduce both network complexity and over-fitting \citep{han2015learning}.
By discarding a fraction $\varsigma$ of the model parameters, the computational complexity may be reduced by a factor of $1-\varsigma$. However, eliminating a substantial fraction of the parameters reduces
the ability of the network to take advantage of 
detailed
correlations in the data.  

Early studies \citep{248452} consider two main types of pruning,
namely sensitivity and penalty-term methods. Sensitivity methods rely on the estimation of influence of a specific node or weight. Penalty-term methods modify the objective function
to force neural networks to remove redundant weights during training.
Although some methods combine both approaches \citep{finnoff1993improving}, others cannot be easily added to
the other family of methods \citep{whitley1990evolution}.

Pruning algorithms are sometimes divided into the following
broad categories: weight pruning vs.\ neuron pruning \citep{li2018optimization}. 
The distinction is in the fact that neuron pruning removes entire neurons (columns/rows in weight matrix) whereas weight pruning is applied on individual entries of the weight matrices. 
This terminology is less suited to describe
some recent methods, where entire subnetworks may be pruned;
\eg heads in transformers or filters in convolution layers \citep{voita-etal-2019-analyzing, he2018soft}.

Recent studies therefore divide pruning techniques into \emph{structured} and \emph{unstructured} methods.
Unstructured pruning is usually applied on individual parameters and does not follow a specific pattern or constraint (\eg \cite{lecun1990optimal,han2015learning}), while structured methods keep the network architecture intact and pruning is applied at the level of filters, channels, or layers \citep{wen2016learning, lin2017runtime}.
An argument against unstructured pruning is that obtaining real performance gains seems hard to achieve without dedicated hardware/libraries \citep{han2016eie}.
In turn, some authors have argued that
models obtained by structured pruning seem to be less accurate \citep{vooturi2018hierarchical}.

Finally, there is another strategy on how to use network pruning, quite different from the ideas mentioned above. 
The goal is to train large models which at inference time require
the same computational cost as smaller ones, by pruning
certain computation paths depending on the input. This idea is referred to as  
\emph{dynamic execution} \citep{gao2018dynamic}, \emph{runtime pruning} \citep{lin2017runtime}, or more broadly \emph{conditional computing} \citep{bengio2013deep}, and it is this paradigm that our work follows.

\subsection{Unstructured Network Pruning}
\label{subsec:unstructured}
An early contribution  
of unstructured pruning
was the use of 
hyperbolic and exponential biases to decay network weights \citep{hanson1989comparing}.
The Optimal Brain Damage and Surgeon methods \citep{lecun1990optimal,hassibi1993second} allowed reducing
the number of connections based on the Hessian of the loss function.
\cite{han2015learning} proposed a magnitude-based pruning method, whereby
weights below a user-defined 
threshold were truncated. \cite{narang2017exploring} applied magnitude pruning on weight matrices of Recurrent Neural Networks (RNNs). They pruned weights below a threshold, monotonically increased during training.
As opposed to
previous techniques, where pruned connections were
lost permanently, \cite{guo2016dynamic} proposed a magnitude-based pruning schema in which eliminated connections could be recovered with some probability.
Some studies employed regularization as a proxy to induce sparseness. \cite{louizos2017learning} utilized the $L_0$-norm to learn sparse networks. 
\cite{molchanov2017variational} devised variational dropout to truncate redundant weights. However, these regularization methods usually suffer from the lack of control over the level of sparseness.

\subsection{Structured Network Pruning}
\label{subsec:structured}
Recent research on network pruning has seen several interesting contributions on structured pruning methods.

Regularization-based pruning can be devised in a structured way. \cite{wen2016learning} proposed a framework based on a group-lasso penalty to remove network components in filter-wise, channel-wise, shape-wise, and depth-wise formats. 
Similarly, \cite{wen2017learning} showed how rows and columns in weight matrices of the Long Short Term Memory (LSTM) can be pruned via a group lasso regularization. 
\cite{liu2017learning} induced sparsity on the batch normalization scaling factors to prune channels. They impose $L_{1}$ regularization on the scaling factors in batch norm to push them toward zero in order to identify insignificant channels. \cite{he2018soft} brought ideas from \cite{guo2016dynamic} into structured pruning by introducing soft filter pruning, whereby during training the filters could recover after first being pruned.
More recently, \cite{voita-etal-2019-analyzing} proposed a method which eliminates the attention heads in transformers using a gating mechanism and an $L_0$ regularization. 

Neuron pruning, as introduced in the previous \secref{subsec:background-pruning-strategy}, can be considered a form of structured pruning.
\cite{srinivas2015learning} presented a variation of the ReLU activation function that has learnable parameters which multiply with neuron’s output. The parameters are encouraged to take binary values with the help of regularizers that reward values close to 0 or 1. Those neurons with a
zero value can simply be removed.
\cite{hu2016network} proposed to prune
individual neurons based on statistics of the network output.

Structured pruning is sometimes referred to as group pruning, whereby entire groups of elements are pruned at once \citep{neta_zmora_2019_3268730}.
Various approaches exist, but the most popular idea is  block pruning in which entire blocks of weights are removed.
Notable contributions are from \cite{narang2017block}, who applied block-wise pruning on RNNs, and \cite{varma2019dynamic}, who introduced Dynamic Block Sparse Reparameterizations (DBSR) where sparse patterns for convolutional layers are learned with an $L_1$ regularizer. 
However, contrary to our approach, their generated patterns are static at inference time.  
\cite{demeester2018predefined} proposed ideas to enforce block-wise sparseness \emph{up front} in word embedding layers and recurrent networks, to also benefit from sparseness during training. Their block-sparse RNN layers were shown to be equivalent to multiple smaller dense RNNs in parallel, each focusing on sub-regions of the input, with concatenated outputs.

\cite{van2019rethinking} presented a more general exploration of
ways to induce sparseness up front on recurrent architectures. As a result, they observed that some types of sparse RNNs (\eg the DiagonalRNN) offer better parallellization and acceleration possibilities in comparison to standard architectures. 

\subsection{Conditional Computation}\label{subsec:conditional}
Conditional computation has been proposed previously to improve model performance without a proportional increase in computational costs.
\cite{bolukbasi2017adaptive} created a pipeline by stacking multiple DNN models and designed a decision function for what they called `early-exit'. Its purpose was to decide for each input, which of the intermediate models would already allow for a correct prediction.
In effect, the system avoided the computational time associated with full evaluation of the pipeline.
\cite{almahairi2016dynamic} introduced the so-called dynamic capacity network, whereby it adaptively focused on task-specific regions of the input data. The proposed model consisted of two modules:
a low capacity network that is activated on the whole input to find task-specific regions, and a high capacity network that is directed by an attention mechanism to focus on the selected regions.

Imposing sparseness in a Mixture of Experts (MoE) can also be seen as a form of conditional computation. \cite{shazeer2017outrageously} proposed the gated MoE, consisting of up to thousands of experts, where a trainable mechanism determines a sparse combination of the experts to use for each example. 
The gating mechanism cannot always be represented by differentiable functions and thereby some studies resorted to reinforcement learning techniques. For instance, \cite{lin2017runtime, wu2018blockdrop} trained a policy network with reinforcement learning to choose which filters or residual-blocks to enable for a given input. 
However, these methods usually require significant
computing resources for training.

The studies that are most closely related to our approach are \cite{gao2018dynamic} and \cite {chen2019you}, which applied conditional gating on channels and filters in Convolution Neural Networks (CNNs). However, in our approach, we devise a gating mechanism on a block level and it is applicable for any matrix-vector multiplication (rather than being CNN-specific).

\section{Dynamic Sparse Linear Layer}
\label{sec:model}
In this section, we motivate and explain our proposed block-wise linear layer with dynamic sparseness. It can be applied in any neural network component that contains a matrix-vector product, to speed up computation at inference time.

Consider a neural network layer that requires a matrix-vector product $W\h$ between a weight matrix $W \in \R^{m\times n}$ and a vector $\h\in\R^{n}$, which in general would be the output from a previous layer. 
Our goal is to strongly reduce the computational load of this operation ($m\,n$ multiplications, $m\,(n-1)$ summations) at inference time. This could be achieved by a static sparse counterpart,
mentioned in \secref{sec:background}. Indeed, setting a number of well-chosen entries in $W$ to zero would lead to reduced requirements both in computation and memory. However, the model with the remaining entries in $W$ has to learn a suitable transformation for the entire space of possible inputs $\h$, which may become harder when fewer parameters in $W$ remain available to the model.
Indeed, limiting the number of non-zero entries in $W$ may correspondingly reduce the expressiveness of the model. The underlying idea behind dynamic sparseness is to decide at run-time which entries in $W$ can be ignored, based on $\h$. Dynamic sparseness therefore stems from the pursuit of achieving a similar gain in computations at inference time, while still

having the flexibility of training all individual weight matrix entries.

\subsection{Sparseness through Gating}

A sparse matrix can be seen as the result of applying a binary masking matrix $G$ to a dense matrix $W$ in an element-wise product. This mechanism is also called gating: entry $W_{ij}$ is zeroed out when the corresponding gate is closed ($G_{ij}=0$), and remains unchanged when the gate is open ($G_{ij}=1$).

More generally, gating can be 
categorized into \emph{continuous} and \emph{discrete} gating. 
On the one hand, continuous forms of gating such as soft attention \citep{bahdanau2014neural} and gated linear units \citep{dauphin2017language} have desirable properties which allow \emph{learning} the gating coefficients,
but they do not offer actual sparseness. For example, when calculating gating coefficients with a sigmoid function, closed gates correspond to very small values, which are never exactly zero, though.
Discrete gating, on the other hand, where closed gates correspond to $G_{ij}=0$,
effectively achieve sparse operations.
For instance, the static sparseness where only selected entries in $W$ are non-zero, can be seen as a predetermined form of discrete gating. 

In the following paragraphs, we propose a gating mechanism which combines the advantages of both: it leads to sparse operations, but is trainable with gradient-based methods as in the continuous gating case. Note that other techniques than the one proposed here are possible, and we discuss some of these in \secref{subsec:alternat_gating}.

In order to make the gating mask applied to $W$ dependent on the vector $\h$,
we need a decision function $\mathcal{G}(\h, \theta, \varsigma)\colon\R^{n}\mapsto\R^{m\times n}$ which maps $\h$ to an $m\times n$ gating mask. The function~$\mathcal{G}$ should be able to meaningfully differentiate between individual $\h$, and it should be trainable in terms of its parameters $\theta$. Furthermore, for a user-specified sparseness level $\varsigma$, a fraction $\varsigma$ of its entries should be zero. It is defined as follows:

\begin{equation}
\mathcal{G}(\h, \theta, \varsigma) = \frac{\topk_{\varsigma}\Big(\FF(\h,\theta)\Big)}{\bigg\langle\topk_{\varsigma}\Big(\FF(\h,\theta)\Big)\bigg\rangle}.\label{eq:gating}
\end{equation}

The function $\FF(\h, \theta)\colon\R^{n} \mapsto \R^{m\times n}$ represents a feed-forward neural network with trainable parameters $\theta$. In our experiments we use a single layer, with $\text{ReLU}$ activation. The $m\,n$ output activations of $\FF(\h,\theta)$ are organized in an $m$ by $n$ masking matrix.
The function $\topk_{\varsigma}$ induces sparseness by retaining the top $k$ activations and setting all others to zero, where $k$ corresponds to the required number of non-zero entries, \ie $k=(1-\varsigma)\,m\,n$ (rounded to the nearest integer). 
Each entry in the resulting matrix is subsequently divided by its mean value, denoted by the denominator in \equref{eq:gating}. With this normalization, the mean value of the gates is always $1$. 
As such, for any distribution of scores in the gating mask, or when changing $\varsigma$ during training (see \secref{subsec:results}), the net `mass' of applied gating always remains the same. 
Our experimental results are in line with this intuition: the normalization is needed for robustness during training and leads to better results.

Finally, the dynamic sparse matrix-vector product is obtained as $\big(\mathcal{G}(\h, \theta, \varsigma)\odot W\big)\ \h$. The general formulation outlined above is however computationally heavier than the original product $W\,\h$. The next paragraphs outline our suggested simplifications of \equref{eq:gating} to arrive at an overall more efficient model.

\subsection{Modifications for Efficiency}
\label{subsec:efficiency}

The gating mechanism outlined above introduces extra parameters and computational complexity. We can reduce that overhead in practice by using block-wise sparse operations, which limits the output size of the feed-forward layer $\FF(\h,\theta)$, and can be made considerably more efficient than other sparse representation methods such as the Compressed Sparse Row (CSR) format \citep{aktulga2014optimizing}.
Note that the block-wise implementation limits the granularity on the sparseness level that can be achieved, but compared to unstructured sparseness
it allows avoiding issues 
related to indexing storage overhead and irregular memory accesses \citep{narang2017block}. 

In particular, we partition the row-dimension of $W$ into $r$ segments, and its column-dimension into $c$ segments. We can then write $W$ as a block matrix
\begin{align*}
W = 
\begin{pmatrix}
W_{11} & \cdots & W_{1c} \\
\vdots & \ddots & \vdots \\
W_{r1} & \cdots & W_{rc} \\
\end{pmatrix}
\end{align*}
and assign a single gate value to each submatrix $W_{ij}$, rather than to each individual scalar entry. We assume that while training, the neural network is capable of organizing its hidden state dimensions in a way that permits the simultaneous switching on or off of these entire submatrices through the gating. The output dimension of the feed-forward layer $\FF$ hence reduces from $n\,m$ to $r\,c$. 

We can go even further, and assume that the network should be able to organize itself if only part of $\h$ is actually used as the key to calculate the gating coefficients. As a result, the input dimension of $\FF(\h, \theta)$ can be reduced as well.

The gating mechanism becomes potentially much cheaper thanks to the ideas of
\begin{enumerate*}[(i)]
\item using a coarser grid of gate coefficients in combination with $W$ as a block matrix, and 
\item using a specific subregion from the $\h$ space to calculate these gates. 
\end{enumerate*}
In our experimental results in \secref{subsec:results} we will quantify the resulting computational gain.

\subsection{Alternatives for Dynamic Gating}\label{subsec:alternat_gating}
As mentioned above, the top $k$ selection of scores through $\topk_{\varsigma}$ as non-zero gate values is our method of choice, because it is fast, and yields good results with gradient based optimization techniques.
Alternatively, dynamic gating could be achieved by sigmoid functions, as in recurrent networks such as an LSTM. This would however leave little control over the sparseness level. Using a softmax function
might constitute another alternative. However, the softmax would focus on a single gating coefficient. This can be circumvented by using the sum of as many softmaxes as one wants non-zero gating values:
\begin{equation}
 G(\h, \theta, \varsigma) = \frac{1}{1-\varsigma}\sum_{i=1}^{k}{\softmax(\FF_i(\h,\theta_i})/\tau),
 \label{eq:sm}
\end{equation}
in which the $\FF_i$ $(i=1\ldots k)$ are feed-forward layers tuned to steer the gating of each softmax component. The parameter $\tau$ is a temperature value. By gradually decreasing it, each softmax becomes more peaky and learns to selects a single entry. At inference time, at most $k$ gates would open, corresponding to a sparseness level of at least $(1-\varsigma)\,r\,c$ (for the block-wise implementation similar to \secref{subsec:efficiency}). Note the normalization in \equref{eq:sm} leads to mean gating values of $1$, similar to \equref{eq:gating}.
While this approach appeared to actually work for small cases, it is not computationally feasible for large-scale problems.

\section{Experimental Results}
\label{sec:experiments}
In this section, we present our experimental results.
We will start by visualizing gate coefficients for a simple feed-forward network (\secref{sec:illustration_on_mnist}). This serves as a sanity check that our method is trainable, and does not lead to the collapsing of the open gates to the same positions for any input.
After that, 
we will evaluate the presence of block-wise dynamic layers in neural sequence models for language modeling 
(\secref{subsec:results}), and provide further insights on dynamic gate distributions. 

\subsection{Illustration on MNIST}
\label{sec:illustration_on_mnist}
\label{sec:experiment:MNIST}
We first demonstrate the dynamical sparseness model on a simple model 
for the task of digit recognition on the MNIST dataset \citep{lecun1998mnist}. 
Although CNN-based architectures such as AlexNet \citep{krizhevsky2012imagenet} would reach higher accuracies, our goal is mainly to apply our ideas to the most basic model, and visualize the gating mechanism. 

We therefore chose for a simple feed-forward neural network, with 5 dynamic sparse hidden layers (each with 1024 units, and block size of 128$\times$128), with sparseness level $\varsigma=90\%$. The sparseness level was kept constant during the whole training process.

We trained the model for 45 epochs using momentum SGD, with a batch size of 128, and a learning rate of \num{5e-3}. It is worth noting that using an overly high learning rate causes drastic updates on the gating mechanism, which leads the model to only focus on a few blocks.  

\begin{figure}
  \centering
  \includegraphics[width=1.0\columnwidth]{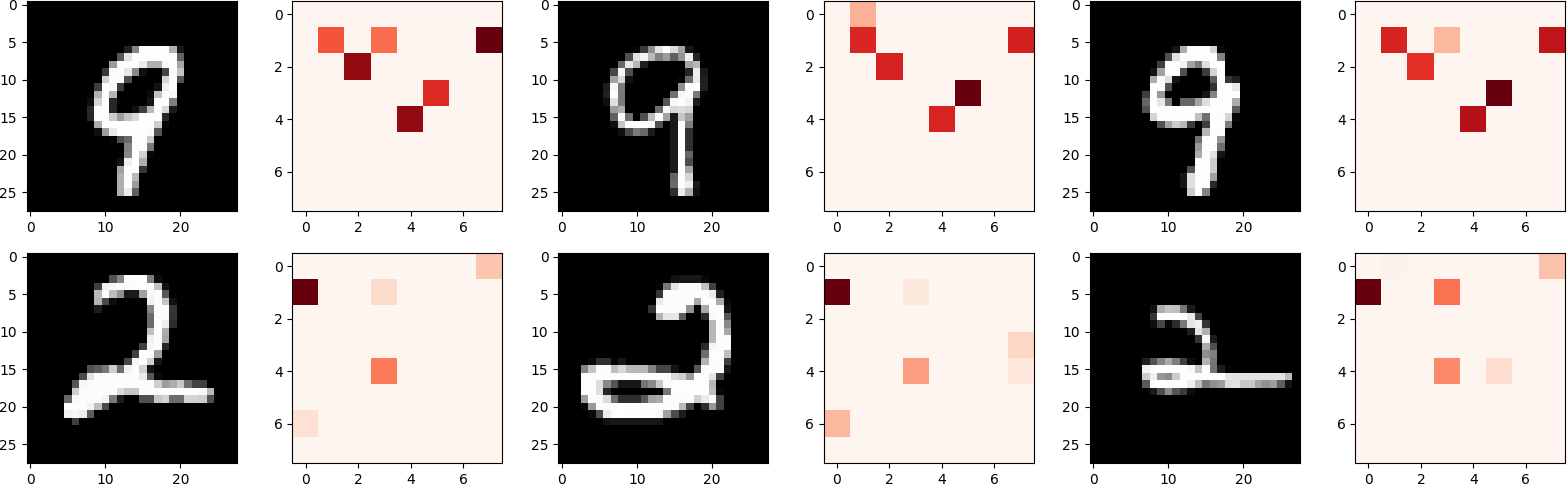}
  \caption{Block specialization in the last hidden layer of a feed-forward network. The generated heat-map demonstrates that the model selects similar patterns for given instances of the same class.}
  \label{fig:path_selection}
\end{figure}

We observe that after around 21 epochs the test error stabilizes to 2\%, which is comparable to the dense case. Moreover, to validate our initial hypothesis that a dynamic sparseness model can 
benefit from addressing different blocks for distinct inputs, 
we looked into the last hidden layer of the network for two different classes. As shown in \figref{fig:path_selection}, for each class, a particular pattern specialization can be perceived. The dynamic gating not only employs
an almost identical pattern of non-zero blocks for instances of the same class,
but also the values are correlated among those same-class inputs (cf.\ similar color intensities).
These results suggest that the dynamic sparsity model we applied has sufficient capacity to act in place of a dense one.

\subsection{Language Modeling Experiments}
\label{subsec:results}
\label{sec:experiments:LM}
In this section we consider the task of language modeling with recurrent sequence models, on two standard datasets, \ie the Penn TreeBank (PTB) from \cite{marcus1993building} and Wikitext2 from \cite{merity2016pointer}. 
The PTB has a vocabulary size of 10,000 words, and contains 1,036,580 tokens, whereas Wikitext2 has a vocabulary size of 33,278, for a total of 2,551,843 tokens.
We apply our proposed model on two well-known architectures, the Quasi-RNN \citep{bradbury2016quasi} and the regularized LSTM \citep{zaremba2014recurrent}.

In the following experiments, we maintain the original dense word embedding layer for these models, since the vocabulary size is relatively small and selecting a vector from a matrix is already computationally efficient. 

For the language modeling experiments, the dynamic sparseness was introduced by increasing $\varsigma$ linearly from zero to the reported final level, over a limited number of epochs during training. We noticed that this yields slightly better results compared to training with a fixed $\varsigma$ from the start, as was done for the MNIST experiments.

\subsubsection{Quasi-RNN Language Model}
A Quasi-RNN layer alternates convolutional layers, parallel across timesteps, and a recurrent pooling function, parallel across channels. As a result it is more time-efficient than fully recurrent models. 
We use a 4-layer Quasi-RNN, with embedding size of 512,  hidden size $\vert\h\vert= 1536$, and tied weights for encoder/decoder. 
Note that the dimensions are slightly different from the original paper (embedding size $400$, hidden size $1550$), because our implementation assumes powers of 2 for the dimensions of the blocks in the block-wise dynamic sparseness model. 
We use the Adam optimizer \citep{kingma2014adam} with learning rate of 0.001, and linearly raise the target sparseness level $\varsigma$ up to the intended value between training epochs 350 and 450.
Otherwise we keep the training setup from \cite{bradbury2016quasi}. For reasons of fair comparison, we do not perform any additional hyper-parameter tuning for our models.\footnote{We use an existing implementation of the Quasi-RNN from fast.ai, adapted for pytorch 1.0.0}

\paragraph{Language modeling effectiveness}
We now investigate the impact of our dynamic sparseness on the language modeling effectiveness for PTB.  In particular, we apply our model on the filters of the convolutional layers. The results are shown in \tabref{tab:T1}, for varying sparseness levels. 
We compare our method (`dyn.~sparse') with the original model (`orig.~dense') as well as a dense model with reduced $\vert\h\vert$ (`small dense') which requires the same number of multiplication-addition operations at inference time as our method. Note that we do not tune any hyper-parameters for that baseline either.  For reference, \tabref{tab:T1} also lists the results from \cite{bradbury2016quasi} as `orig.\ ref'. The number of parameters (`params') in their model is 3.6M lower than the `orig.\ dense' model, because of the lower dimension for embedding layers (see above). Also, their perplexity is slightly higher than our dense baseline, which may be related to our different implementation with another optimizer.  
Besides $\vert\h\vert$ and the perplexity (`ppl'), the table lists the sparseness level $\varsigma$, and the fraction of multiplication-addition computations in the matrix-vector product at inference time (denoted as `comput.'), relative to the dense model. 

Introducing sparsity, \ie reducing the number of matrix-vector products,
leads to increased perplexities, both for the dynamic sparse model and for the small dense baseline.
However, the \emph{dynamic} sparse models consistently outperform the \emph{small dense} baselines with corresponding computational cost.
For example, our technique allows halving the number of multiplication-addition operations (comput.\ $=0.5$) at the cost of only $+1.2$ perplexity points, compared to 
$+2.3$ for the smaller dense model at the same computational cost. 

Note that the mentioned increase in perplexity in going from dense to $50\%$ sparse is an upper bound, and may be related to the additional regularization effect of the dynamic sparseness. This effect is not compensated for, as the dropout probabilities are not tuned. Indeed, we noticed that switching off weight drop regularization in both models, leads to the same (yet, higher) perplexity in the dense and the $50\%$ sparse model. 

We observe a small increase in the number of parameters, from $47.4$M to $47.9$M, for the calculation of the gate coefficients. Yet, in our experiments, the amount of corresponding additional 
multiplication-addition operations is an order of magnitude lower than the total computational gain that can be achieved by dynamic sparseness.

\begin{table}[t!]
\centering
\begin{threeparttable}
\caption{Language modeling perplexity (ppl) of the reference implementation (orig.~ref) from \cite{bradbury2016quasi}, original Quasi-RNN (orig.~dense), the small dense, and the dynamic sparse model, on the PTB test set. Column `comput.' denotes the fraction of matrix-vector product computations compared to the dense baseline.
}
\begin{tabular}{lcccccc@{}} 

\toprule
Model & $\vert \h\vert$ & $\varsigma$ & ppl & comput. & params \\
\midrule
orig.~ref & $1550$ & $0\%$ & $56.70$ & - & $43.8$M \\
orig.~dense                  & $1536$   &  $0\%$  &  $54.93$  & $1.0$ & $47.4$M  \\
\midrule
               
                                      & $810$  & $0\%$     & $57.23$  & $0.5$ & $20.3$M \\
small dense                            & $660$  & $0\%$     & $58.76$  & $0.4$ & $16.3$M \\
& $500$  & $0\%$     & $60.82$  & $0.2$  & $12.6$M \\
                                      & $350$  & $0\%$     & $64.08$  & $0.1$  & $9.7$M \\
\midrule
               
                            & $1536$  & $50\%$     & $56.17$  & $0.5$ & $47.9$M \\
dyn.~sparse                        & $1536$  & $60\%$     & $57.93$  & $0.4$ & $47.9$M \\
(ours)                            & $1536$  & $80\%$     & $60.70$  & $0.2$ & $47.9$M \\
                            & $1536$  & $90\%$     & $63.59$  & $0.1$ & $47.9$M \\
\bottomrule

\end{tabular}
\label{tab:T1}
\end{threeparttable}
\end{table}

\begin{figure}
  \centering
  \includegraphics[scale=0.3]{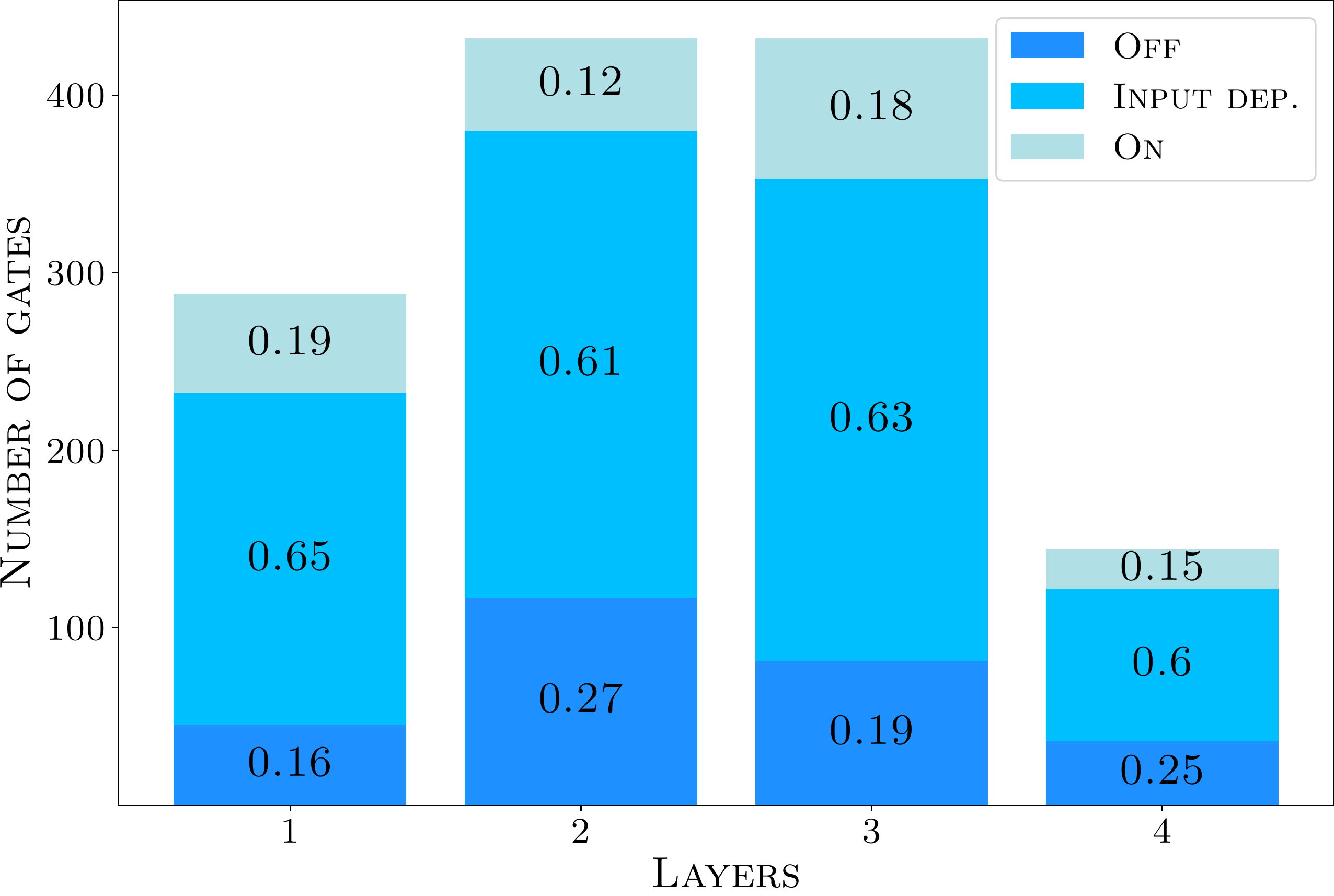}
  \caption{Gate distribution on PTB test set: number of gates switched off or on over $95\%$ of the time (\textsc{Off} and \textsc{On}, resp.) or in between (\textsc{Input dep.}), for the various Quasi-RNN layers. Corresponding fractions are shown on the bars.}
  \label{fig:gate_distribution}
\end{figure}

\paragraph{Analysis of the gating mechanism}
We expect three different ways the gating mechanism may function in a network with dynamic computations: 
\begin{enumerate*}[(i)]
\item \label{step:1} Some gates will be always active: we expect certain blocks in a network to be of key importance for all types of inputs;
\item \label{step:2} Other gates would become active conditioned on the input: the blocks in the weight matrices that are more specialized for certain features are dynamically selected based on the input;
\item \label{step:3} Finally, some gates may always remain closed, corresponding to static sparsity.
\end{enumerate*} 
We categorized gates as `always' on/off, if they are on/off for more than 95\% of the instances on the test set.
\Figref{fig:gate_distribution} shows how the gates of the Quasi-RNN model with $\varsigma=0.50$ are distributed among these categories, for each of the 4 Quasi-RNN layers. 
The total height of each bar indicates the total number of gates for the corresponding layer (with smaller numbers of gates for the first and last layer, as the embedding size is smaller than the hidden state size). 
We observe that at least $60\%$ of the gates in all layers are actually input-dependent.
This suggests that the model prefers dynamic sparsity over static sparsity. 

\subsubsection{LSTM Language Model}
We now train an LSTM language model on the PTB and Wikitext2 datasets, using a model and training procedure similar to the one described in \cite{zaremba2014recurrent}.
The model is composed of an embedding layer, 2 LSTM layers, and a softmax layer. 
Each LSTM layer has 1536 hidden units, and a $65\%$ dropout rate is applied on the non-recurrent connections. The dimension of word embeddings in the input layer is 1536.
We train the model for 55 epochs and start to increase the sparseness level from epoch 25 to 40 to reach the predefined level. Dynamic sparseness was applied to the input-to-hidden and hidden-to-hidden weight matrices in the LSTMs. 
Besides the original dense baseline, we also apply the static sparsity model called Automated Gradual Pruning (AGP) \citep{zhu2017prune}, which gradually prunes parameters based on their weight magnitude.\footnote{We used the existing implementation provided in the Distiller library \citep{neta_zmora_2019_3268730} for the AGP method}

\Tabref{tab:T2} summarizes the results. There is a general degradation in the model quality with increasing sparseness levels, although with our $50\%$ dynamic sparse model the baseline perplexity is still maintained. 
Our dynamic model consistently outperforms the static sparse model. This confirms our hypothesis that dynamic sparseness retains more expressiveness than static sparseness. If the computational cost at inference is more critical than memory, dynamic sparsity therefore presents a valid alternative. 
We further noticed that compensating for the regularization effect of the dynamic sparseness by slightly lowering the dropout rate compared to the dense case may lead to improved results for the block-wise dynamic sparseness model. 
Tuning the dropout rate over the values $\{0.4, 0.50, 0.55, 0.60, 0.65\}$ resulted in a $1.31$ perplexity point decrement on PTB and $0.57$ on wikitext-2 for the dynamic sparse model (with $0.60$ as optimal dropout rate). However, similar tuning for the AGP method did not lead to further improvements. 

\begin{table}[t!]
\centering
 \caption{LSTM language model test perplexity on PTB and Wikitext2. Comparing a standard dense LSTM model with a static sparse model and our dynamic sparse model, for different sparseness levels $\varsigma$.}
\begin{tabular}{lccccc}
\toprule
Model & $\varsigma$ & PTB & Wikitext2 \\
\midrule

 \cite{zaremba2014recurrent} & $0\%$  & $78.40$ & - \\
orig.~dense                  & $0\%$  &  $78.57$  & $85.0$  \\
\midrule
               
static sparse (AGP)                           & $50\%$     & $80.92$  & $86.91$  \\
                              & $75\%$     & $81.06$  & $88.45$  \\
                              & $90\%$     & $81.75$  & $89.48$  \\
\midrule
               
dyn.~sparse (ours)                     & $50\%$     & $78.50$  & $85.31$  \\
                              & $75\%$     & $79.12$  & $86.76$  \\
                              & $90\%$     & $80.34$  & $88.44$  \\
                            
\bottomrule
\end{tabular}

\label{tab:T2}
\end{table}

\section{Conclusion}
\label{sec:conclusion}
We proposed the technique of block-wise dynamic sparseness, which can be used to reduce the computational cost at inference time for matrix vector products inside neural network building blocks.
Our experimental results on Quasi-RNN and LSTM based language models provide a proof-of-concept of the proposed method, with significant reduction of the required computations at inference time at a very limited model effectiveness penalty.
Under the same experimental settings,  
our method outperforms a baseline with static sparseness. 
Finally, the implementation of a cuda kernel to support the dynamic gating, and our code to reproduce the presented experimental results, are made publicly available
.\footnote{https://github.com/hadifar/dynamic-sparseness}

\bibliographystyle{model2-names}
\bibliography{references}

\end{document}